\documentclass[conference]{IEEEtran}
\IEEEoverridecommandlockouts
\usepackage{cite}
\usepackage{amsmath,amssymb,amsfonts}
\usepackage{algorithmic}
\usepackage{graphicx}
\usepackage{textcomp}
\usepackage{xcolor}
\usepackage[raggedright]{sidecap}
\usepackage[singlelinecheck=false,justification=justified]{caption}
\usepackage{booktabs}
\usepackage{multirow}
\usepackage{enumitem}
\usepackage{float}
\usepackage[table]{xcolor}
\usepackage[utf8]{inputenc}
\usepackage{listings}

\usepackage{caption}
\usepackage{subcaption}
\usepackage{balance}
\usepackage{sidecap}
\sidecaptionvpos{figure}{t}
\usepackage[colorlinks]{hyperref}
\usepackage{cite}
\newcommand\blfootnote[1]{%
  \begingroup
  \renewcommand\thefootnote{}\footnote{#1}%
  \addtocounter{footnote}{-1}%
  \endgroup
}

\def\BibTeX{{\rm B\kern-.05em{\sc i\kern-.025em b}\kern-.08em
    T\kern-.1667em\lower.7ex\hbox{E}\kern-.125emX}}
\begin{document}

\title{POLY-SIM: Polyglot Speaker Identification with Missing Modality Grand Challenge 2026 \\Evaluation Plan}


\author{Marta Moscati$^{1}$\textsuperscript{\textdagger}, Muhammad Saad Saeed$^{2}$\textsuperscript{\textdagger}, Marina Zanoni$^{1,3}$, Mubashir Noman$^{4}$, Rohan Kumar Das$^{5}$,\\ Monorama Swain$^{1}$, Yufang Hou$^{6}$, Elisabeth André$^{7}$, Khalid Mahmood Malik$^{1}$, Markus Schedl$^{1,8}$ Shah Nawaz$^{1}$  \\
$^{1}$Institute of Computational Perception, Johannes Kepler University Linz, Austria, \\
$^{2}$University of Michigan-Flint, USA,
$^{3}$Sapienza University of Rome, Italy\\
$^{4}$Mohamed bin Zayed University of Artificial Intelligence,\\
$^{5}$Fortemedia Singapore, Singapore, 
$^{6}$IT:U Interdisciplinary Transformation University Austria, \\
$^{7}$University of Augsburg, Germany,
$^{8}$Human-centered AI Group, AI Lab, Linz Institute of Technology, Austria \\
\tt mavceleb@gmail.com
}

\maketitle

\begin{abstract}
Multimodal speaker identification systems typically assume the availability of complete and homogeneous audio–visual modalities during both training and testing. However, in real-world applications, such assumptions often do not hold. Visual information may be missing due to occlusions, camera failures, or privacy constraints, while multilingual speakers introduce additional complexity due to linguistic variability across languages. These challenges significantly affect the robustness and generalization of multimodal speaker identification systems.
The POLY-SIM Grand Challenge 2026 aims to advance research in multimodal speaker identification under missing-modality and cross-lingual conditions. Specifically, the Grand Challenge encourages the development of robust methods that can effectively leverage incomplete multimodal inputs while maintaining strong performance across different languages.
This report presents the design and organization of the POLY-SIM Grand Challenge 2026, including the dataset, task formulation, evaluation protocol, and baseline model. By providing a standardized benchmark and evaluation framework, the challenge aims to foster progress toward more robust and practical multimodal speaker identification systems.

\end{abstract}



\begin{NoHyper}
\blfootnote{\textsuperscript{\textdagger}Equal Contribution.}
\end{NoHyper}

\section{Introduction}
\label{sec:intro}
\begin{figure}[!t]
    \centering
    \includegraphics[width=0.95\linewidth]{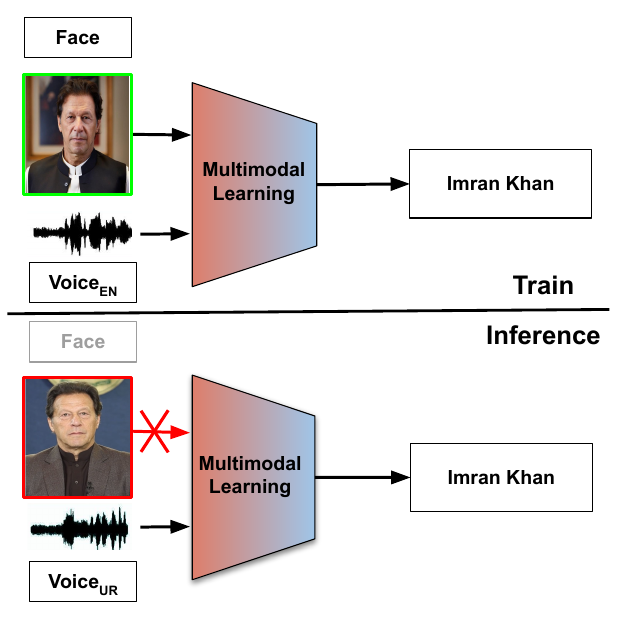}
    \caption{POLY-SIM: Polyglot Speaker Identification with Missing Modality. 
    The model is trained on paired face images and audio segments in a specific language (e.g., English). At test time, the face modality is missing and the input consists of audio segments in another language (e.g., Urdu) only.
    }
    \label{fig:task}
\end{figure}

The face and voice of a person have unique characteristics and they are often used as biometric measures for speaker identification, either as a unimodal or multimodal task~\cite{jain2004introduction}.
Recent advancements have been fueled by the curation of large-scale audio-visual datasets such as VoxCeleb~\cite{nagrani2017voxceleb,chung18b_interspeech,nagrani2020voxceleb} and VoxBlink~\cite{lin24j_interspeech}. These datasets enable the development of multimodal models~\cite{tao20b_interspeech,tao2021someone,jiang23c_interspeech,praveen2025lavvit} for 
speaker identification. 
However, a major limitation of existing models is that they require complete audio-visual modalities to achieve good performance, and consequently experience performance deterioration when modalities are not complete. This issue, referred to as missing-modality, is a well-known challenge in multimodal learning across several tasks~\cite{ma2022multimodal,lin2023missmodal,saeed2024modality,guo2024multimodal,ganhor2024multimodal,ganhor2025single,liaqat2025multimodal,breiteneder2026robust}.
A separate challenge, specific to tasks such as speaker identification, is that of language shifts: when trained on one language (e.g., English) and evaluated on another (e.g., German), existing methods experience a performance deterioration, often attributed to the acoustic and phonetic differences between the train and evaluation languages.
To address these limitations, we propose POLY-SIM, a grand challenge that investigates multimodal learning under missing-modality and cross-lingual settings, as illustrated in Figure~\ref{fig:task}. The challenge is designed to evaluate how well models can leverage partial multimodal inputs while maintaining strong cross-lingual generalization.

\section{Grand Challenge Objective}
\begin{itemize}
    \item Multimodal learning tasks under missing modalities, such as identifying a speaker when visual inputs are unavailable, or when language conditions differ, reflect \textbf{real-world scenarios} faced by modern multimedia systems. Addressing this problem is therefore critical for building robust, flexible, and fair multimedia systems.
    \item For the \textbf{research} community, this challenge pushes advances in representation learning, cross-modal alignment, domain adaptation, and generalization under distribution shift. 
    For \textbf{industry}, it directly impacts the reliability of biometric systems, media analytics, human–computer interaction, and security applications deployed in real-world scenarios.
    \item This challenge is a \textbf{continuation} of the previous FAME 2024~\cite{saeed2024synopsis} and 2026~\cite{moscati2025linking} Grand Challenges, which focused on understanding and analyzing the impact of language on face-voice association. Building on this foundation, the current challenge addresses a critical and increasingly realistic setting in which multimedia models must operate across languages while coping with missing  modalities.
\end{itemize}

\section{Grand Challenge Description}

\subsection{Challenge Setup}
In the grand challenge, a multimodal model is trained on paired face images and segments of speech in one language (e.g., English).
At test time, the face modality is missing and the available speech segment is in a different language (e.g., Urdu), see Figure~\ref{fig:task}. This simulates both (i) the missing visual modality and (ii) the cross-lingual scenario for multimodal speaker identification under real-world scenarios.
Let $\mathcal{D}_{\mathrm{train}}=\{(F_i^{f}, V_i^{a,\ell_{\mathrm{en}}}, y_i)\}_{i=1}^{N}$ represent the training dataset consisting of $N$ samples, where $\ell_{\mathrm{en}}$ is a label indicating the training language, while $F_i^{f}$ and $V_i^{a,\ell_{\mathrm{en}}}$ denote the face and audio modalities, respectively. Each instance is associated with a class label $y_i \in \mathcal{Y}_{i=1}^{S}$, where $S$ denotes the number of speakers. The modality-specific embeddings are defined as $x_i^{f} = \phi_f(F_i^{f})$ and
$x_i^{a,\ell_{\mathrm{en}}} = \phi_a(V_i^{a,\ell_{\mathrm{en}}})$,
where $\phi_f(\cdot)$ and $\phi_a(\cdot)$ denote the face and audio encoders.
At test time, the face modality is missing and only the audio modality is
available in a different language $\ell \in \mathcal{L}_{\mathrm{test}} = \{\ell_{\mathrm{ur}}\}$, 
corresponding to Urdu. The dev or eval set is thus given by $\mathcal{D}_{\mathrm{test}}=\{(V_j^{a,\ell})\}_{j=1}^{M}$;
analogously to the training phase, the audio embedding at inference time is computed as $x_j^{a,\ell} = \phi_a(V_j^{a,\ell})$.
The task's goal is to predict the speaker label $y_j$ based on the audio embedding in language $\ell \neq \ell^\text{en}$, and using a model $f$ trained on paired
audio--visual English data:
\[
\hat{y}_j = f(x_j^{a,\ell}),
\]
despite the \emph{missing-image} and the \emph{language shift} at inference.

\subsection{Evaluation Protocol}
We follow the protocol below to investigate the robustness of multimodal networks under missing-modality and cross-lingual scenarios. 
\begin{enumerate}[label=P\arabic*., start=3]
    \item \textbf{In-language multimodal.} Training and testing on the same language with both modalities available.
    \item \textbf{Missing-modality.} Testing with only the audio modality while the face modality is missing.
    \item \textbf{Cross-lingual and multimodal.} Training on one language and testing on another with both modalities available.
    \item \textbf{Cross-lingual and missing-modality.} Cross-lingual testing under missing-modality.
\end{enumerate}

\begin{figure}[t]
    \centering
    \includegraphics[width=0.95\linewidth]{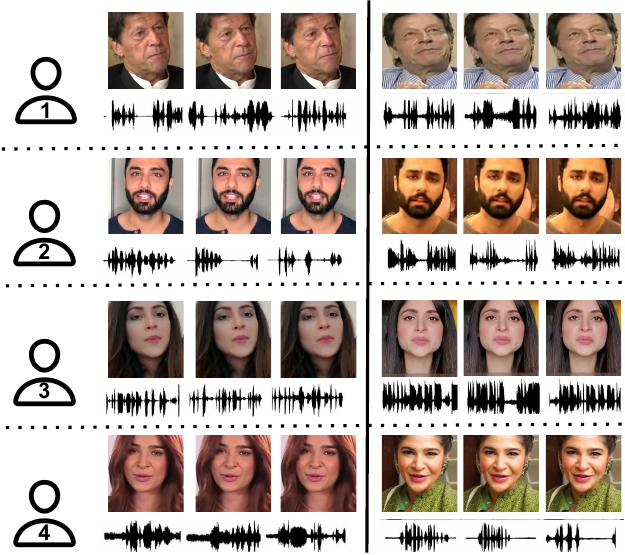}
    \caption{Audio-visual samples randomly selected from the MAV-Celeb~\cite{nawaz2021cross,saeed2024synopsis,moscati2025linking}. The visual data contains different variations such as pose, lighting condition, and motion. (Left) The block shows data of speakers speaking English. (Right) The block shows data of the same speakers speaking Urdu language.}
    \label{fig:mavceleb}
\end{figure}

\begin{table}
\caption{MAV-Celeb dataset statistics for English--Urdu language pair.}
\centering
\renewcommand{\arraystretch}{1.15}
\setlength{\tabcolsep}{10pt}
\scalebox{0.99}{
\begin{tabular}{c|c|c|c}
\toprule
\multirow{2}{*}{\textbf{Lang. Pair}} & \multirow{2}{*}{\textbf{Lang.}} & \textbf{Total Videos} & \textbf{Samples} \\
 &  & \textbf{(Tr./Val./Test)} & \textbf{(Tr./Val./Test)} \\
\midrule
\multirow{2}{*}{Eng--Urdu}
 & Eng & 262 / 70 / 70 & 4039 / 1290 / 1521 \\
 & Urdu    & 415 / 70 / 70 & 9304 / 1779 / 1623 \\
\bottomrule
\end{tabular}
}
\label{tab:dataset_stats}
\end{table}

\begin{table*}[!t]
\centering
\caption{Performance in accuracy (\%) for English--Urdu cross-lingual experiments.}
\renewcommand{\arraystretch}{1.15}
\setlength{\tabcolsep}{16pt}
\scalebox{0.95}{
\begin{tabular}{lcccccc}
\toprule
\multirow{2}{*}{\textbf{Configuration}} & \multirow{2}{*}{\textbf{Phase}} &
\textbf{P3} & \textbf{P4} & \textbf{P5} & \textbf{P6} & \multirow{2}{*}{\textbf{Avg.}} \\
\cline{3-6}
& & \multicolumn{1}{c}{Face--Audio (Eng.)} & \multicolumn{1}{c}{ Audio (Eng.)} &
\multicolumn{1}{c}{Face--Audio (Urdu)}     & \multicolumn{1}{c}{ Audio (Urdu)} & \\
\midrule

\multirow{2}{*}{Face-Audio (Eng.)}
& Progress  & 97.44 & 37.75 & 98.48 & 31.70 & 66.34 \\
& Eval      & 98.82 & 52.53 & 98.27 & 43.87 & 73.37 \\



\bottomrule
\end{tabular}
}
\label{tab:v1_multimodal_configs}
\end{table*}


\subsection{Dataset}
We base our Grand Challenge on the MAV-Celeb dataset, which allows studying the impact of languages on face-voice association formulated as cross-modal verification task~\cite{nawaz2021cross,saeed2024fame,moscati2025linking}. 
The dataset consists of audio-visual samples obtained from YouTube videos of speakers appearing in interviews, talk shows, and television debates. 
Most importantly, each speaker is bilingual, and we select the dataset subset in which each speaker appears in videos while speaking English and Urdu. 
We adapted the dataset for the task of multimodal speaker identification under missing-modality and cross-lingual scenarios. 
Table~\ref{tab:dataset_stats} provides detailed statistics of the dataset, while Figure~\ref{fig:mavceleb} presents audio-visual samples from the newly collected dataset split.
The dataset is publicly available and provided alongside pre-extracted features representing the audios and images as encoded with state-of-the-art pre-trained architectures. 
We release all information related to the challenge on the website.\footnote{\href{https://github.com/msaadsaeed/polysim}{https://github.com/msaadsaeed/polysim}}.

The train split of the dataset is structured hierarchically by modality (faces and voices), identity (\texttt{idxxxx}), and language (English and Urdu), where each identity contains multiple video samples with corresponding image (\texttt{.jpg}) and audio (\texttt{.wav}) files.
The progress and evaluation files are stored in CSV format, where each row corresponds to an audio-visual along with key.




\subsection{Baseline Method \& Starter Kit}
To allow participants to benchmark their results, we will release a pretrained instance of a competitive multimodal method for face-voice association task~\cite{saeed2022fusion}. 
The model consists of a two-branch network that takes as input the embeddings of faces and voices. The embeddings to be used as input to the face-encoding branch are obtained using a popular convolutional neural network pre-trained on a large-scale facial recognition dataset~\cite{schroff2015facenet}.  The embeddings to be used as input to the voice-encoding branch are obtained using an audio encoding network for speaker recognition~\cite{desplanques20_interspeech} trained using the language available in the training set. The multimodal model further combines the face and voice embeddings, and is optimized by means of a loss function that imposes orthogonality constraints on the multimodal embeddings of different speakers.
We refer the readers to FOP~\cite{saeed2022fusion} and to the repository of the dataset for more information on prior work on the baseline.

\subsection{Evaluation Metric}
We use standard P-accuracy as the evaluation metric. P-accuracy measures the proportion of test pairs for which the system correctly predicts the matching identity among P candidates. For monolingual pairs (same training and test language), we report P$3$ Acc and P$4$ Acc. For cross-lingual pairs (different training and test language), we report P$5$ Acc and P$6$ Acc. The overall score is the mean across all four configuration.

\subsection{Baseline Results}
Table~\ref{tab:v1_multimodal_configs} presents the baseline results under missing-modality and cross-lingual settings on the MAV-Celeb dataset. The POLY-SIM 2026 grand challenge encourages participants to develop novel approaches that improve the performance of multimodal speaker identification under these challenging conditions.

\subsection{Submission Template}
The grand challenge will be implemented using CodaBench\footnote{\url{https://www.codabench.org/competitions/11283}}. Participants are expected to submit csv files including the \texttt{keys} and class labels in the following format:

\begin{itemize}
    \item \texttt{key,p3,p4}
    \item \texttt{t5M7dziYVY,1,0}
    \item \texttt{RmUYdg2luC,50,0}
    \item \texttt{BvKCMACzXt,20,0}
    \item \dots
    \item \texttt{TB9XrX8A3i,11,11}
\end{itemize}

Participants must submit a ZIP archive containing CSV files, one per language pair. To create the archive, run zip submission.zip *.csv from within the directory containing the submission files (do not zip the folder itself). Files must be named as follows:

\begin{itemize}
\item \texttt{submission\_v1\_<phase>\_English\_English.csv}
\item \texttt{submission\_v1\_<phase>\_English\_Urdu.csv}
\end{itemize}

\noindent Where \texttt{<phase>} is val (dev) or test (eval). 
Each CSV file must contain a header row and one row per test pair. For monolingual files (lang1 == lang2), the required columns are key, p3, and p4. For cross-lingual files (lang1 != lang2), the required columns are key, p5, and p6. The key is the unique identifier for each test pair, and p3/p4/p5/p6 are the predicted identity indices among P candidates.
In the progress phase, each team will be allowed to submit a maximum of $150$ submissions, with a maximum $15$ per day. 
In the evaluation phase, the number of total submission will be limited to $15$. The overall score will be computed as:

\begin{equation}
\text { Overall Score }= \text{Acc}  (\text  {P3  + P4  + P5  + P6 }) / 4
\end{equation}

\subsection{Timeline}

We are planning the challenge timeline with regarding to ACM MM paper submission as follows:

\begin{itemize}
\item Registration Period: 27 March - 10 May 2026
\item Progress Phase: 27 March - 15 May 2026
\item Evaluation Phase: 16 May - 23 May 2026
\item Challenge Results: 25 May 2026
\item Final Paper: 8 June 2026
\end{itemize}
 
\subsection{Registration Process}
The following Google Form will be used to allow participants to register their teams to the challenge\footnote{\url{https://forms.gle/EwmVBiph2QsZ2QRB9}}.

\subsection{Rules for System Development}
We will enforce the following rules for participation:

\begin{itemize}
    \item The participants are required to submit a system description. Teams without system description will be disqualified from the challenge. Teams describing a setup that violates one of the above rules will be disqualified. 
    \item The participants are required to submit a link to a working version of their setup, e.g., on a platform for open-source development such as GitHub. Teams without code submission or with a setup that violates one of the above rules will be disqualified.
    \end{itemize}

\section*{Acknowledgments}
This research was funded in whole or in part by the Austrian Science Fund (FWF): Cluster of Excellence \href{https://www.bilateral-ai.net/home}{\textcolor{blue}{\textit{Bilateral Artificial Intelligence}}} (\url{https://doi.org/10.55776/COE12}), the doc.funds.connect project \href{https://dfc.hcai.at/}{\textcolor{blue}{\textit{Human-Centered Artificial Intelligence}}} (\url{https://doi.org/10.55776/DFH23}), and the PI project \href{https://doi.org/10.55776/P36413}{\textcolor{blue}{\textit{Intent-aware Music Recommender Systems}}} (\url{https://doi.org/10.55776/P36413}).

\balance
\bibliographystyle{IEEEbib}
\bibliography{IEEEbib}
\end{document}